%% file: main.tex
\documentclass[10pt,twocolumn,letterpaper]{article}

\usepackage[pagenumbers]{cvpr} 

\input{preamble}

\definecolor{cvprblue}{rgb}{0.21,0.49,0.74}
\usepackage[pagebackref,breaklinks,colorlinks,allcolors=cvprblue]{hyperref}
\usepackage{multirow}
\usepackage{kotex}
\usepackage{makecell}
\usepackage{wrapfig}
\usepackage{tabularx}
\usepackage{graphicx}

\newcommand\blfootnote[1]{%
  \begingroup
  \renewcommand\thefootnote{}\footnote{#1}%
  \addtocounter{footnote}{-1}%
  \endgroup
}

\title{Improving Geometry in Sparse-View 3DGS via \\ Reprojection-based DoF Separation}

\author{
Yongsung Kim$^{1,}$\thanks{co-author}~~~~~Minjun Park$^{1,}$\footnotemark[1]~~~~~Jooyoung Choi$^{2}$~~~~~Sungroh Yoon$^{1,2,3,}$\thanks{co-author}\\
$^1$Interdisciplinary Program in AI, Seoul National University\\
$^2$ECE, 
$^3$AIIS, ASRI, INMC, ISRC, Seoul National University\\
{\tt\small \{libary753, minjunpark, jy\_choi, sryoon\}@snu.ac.kr}
}

\begin{document}

\twocolumn[{%
\renewcommand\twocolumn[1][]{#1}%
\maketitle
\begin{center}
    \vspace{-1.9em}
    \centering
    \captionsetup{type=figure}
    \includesvg[width=0.95\linewidth]{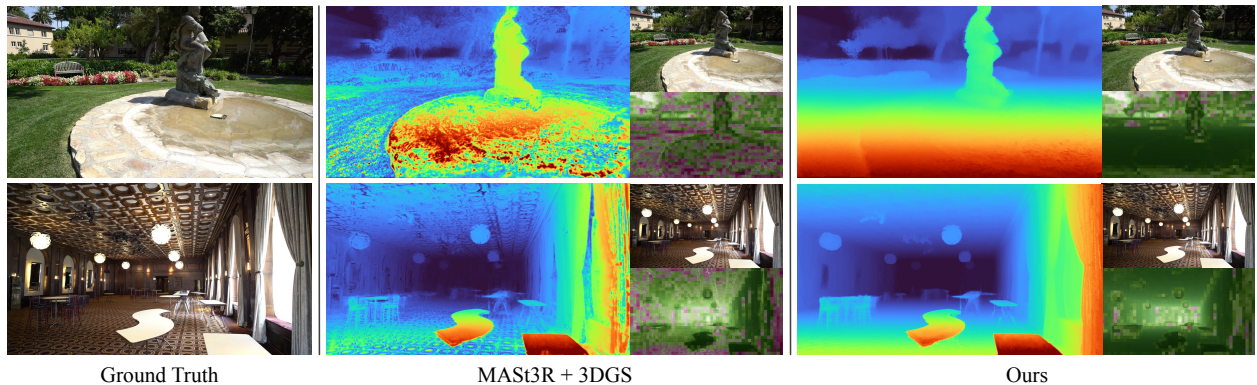}
    \captionof{figure}{
    \textbf{3D reconstruction results with 12 views.} The image on the left side is the rendered depth image, and the upper right image is the rendered RGB image. The lower right image visualizes patch-wise depth correlation, where green indicates accurate geometry. As the color shifts from \textcolor[RGB]{87,156,44}{\textbf{green}} to \textcolor[RGB]{170,170,170}{\textbf{gray}} and then to \textcolor[RGB]{195,31,125}{\textbf{purple}}, the patch-wise depth correlation decreases, indicating less plausible geometry. Our method qualitatively demonstrates more uniform and realistic geometry, which is also evident from the higher patch-wise depth correlation.
    }
    \label{fig:teaser}
\end{center}%
}]


\blfootnote{$\dagger$ Correspondence to: Sungroh Yoon (sryoon@snu.ac.kr)}
\blfootnote{$*$ Both authors contributed equally to this work}

\vspace{-1.1em}

\input{sec/0_abstract}

\input{sec/1_intro}
\input{sec/2_related_work}
\begin{figure}[b]
  \centering
   \includesvg[width=\linewidth]{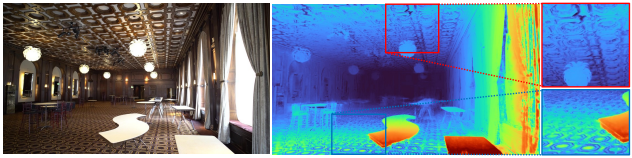}
   \caption{
   \textbf{Geometric artifacts from naive 3DGS refinement.}
   Texture representation via Gaussian positions introduces unintended geometric patterns. The red box highlights excessive distortion in the ceiling geometry, and the blue box shows gaps in the flat floor geometry following texture patterns.
   }
   \label{fig:geom_artifact_example}
\end{figure}
\begin{figure*}[htp]
    \centering
    \includesvg[width=\textwidth]{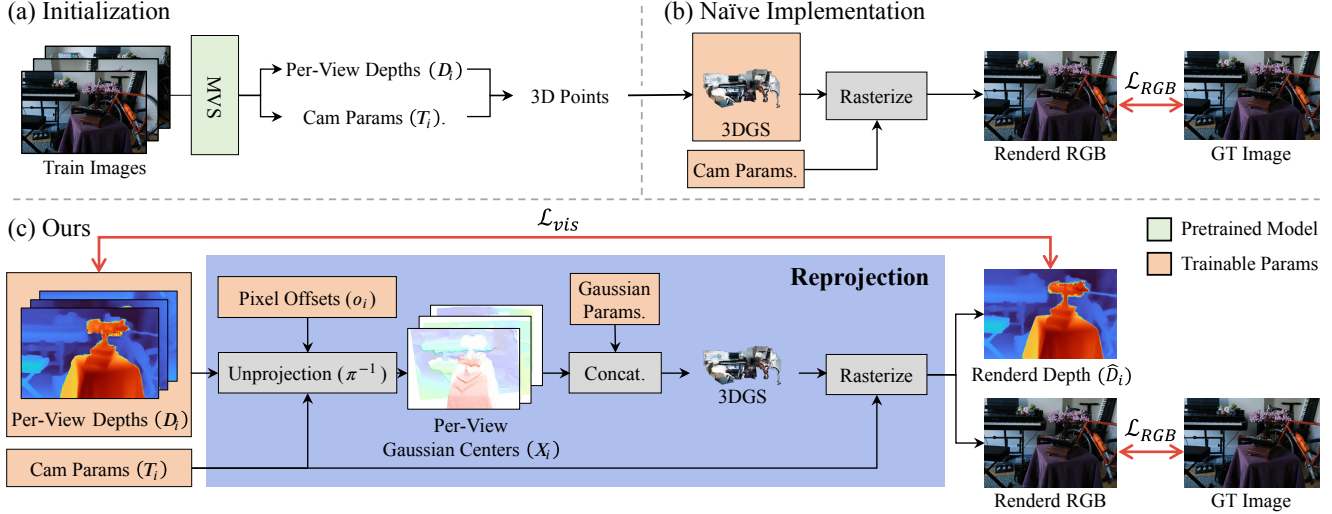}
    \caption{\textbf{Overview of the proposed framework.} (a) Scene initialization using a learning-based MVS model, which predicts 3D points from images and outputs per-view depth as an intermediate representation. (b) Na\"ive implementation, where MVS is treated as a black-box model, and its output is refined using the 3DGS pipeline. (c) Our proposed framework, which introduces reprojection-based refinement by retaining intermediate per-view depth as a trainable target. A visibility loss function is used to resolve conflicts when aligning individual per-view depths into a shared coordinate system.
    }
    \label{fig:overall_framework}
\end{figure*}

\input{sec/3_preliminary}
\input{sec/4_Method}
\input{sec/5_experiment}
\input{sec/6_analysis}

\input{sec/7_summary_and_limiataion}

{
    \small
    \bibliographystyle{ieeenat_fullname}
    \bibliography{main}
}

\end{document}

%% file: preamble.tex
\usepackage{amsmath}
\usepackage[inkscapelatex=false]{svg}
\usepackage{float}

%% file: sec/0_abstract.tex
\begin{abstract}

Recent learning-based Multi-View Stereo models have demonstrated state-of-the-art performance in sparse-view 3D reconstruction. However, directly applying 3D Gaussian Splatting (3DGS) as a refinement step following these models presents challenges. We hypothesize that the excessive positional degrees of freedom (DoFs) in Gaussians induce geometry distortion, fitting color patterns at the cost of structural fidelity.
To address this, we propose reprojection-based DoF separation, a method distinguishing positional DoFs in terms of uncertainty: image-plane-parallel DoFs and ray-aligned DoF.
To independently manage each DoF, we introduce a reprojection process along with tailored constraints for each DoF.
Through experiments across various datasets, we confirm that separating the positional DoFs of Gaussians and applying targeted constraints effectively suppresses geometric artifacts, producing reconstruction results that are both visually and geometrically plausible.

\end{abstract}

%% file: sec/1_intro.tex
\section{Introduction}
\vspace{0.2em}
Advancements in augmented reality, autonomous driving, and robotics have heightened the demand for accurate 3D geometry reconstruction~\cite{geometry_autonomous_drive, ARM_pmlr21}. While various depth sensors exist~\cite{newcombe2011kinectfusion,niessner2013real}, RGB camera-based reconstruction remains a cost-effective and scalable solution~\cite{schonberger2016colmap, mildenhall2020nerf, kerbl20233dgs}. Traditional multi-view stereo (MVS) methods~\cite{furukawa2009accurate,schoenberger2016mvs} depend on robust feature-matching algorithms~\cite{sift, surf} and require a large number of images to retain sufficient 3D features after outlier rejection~\cite{schonberger2016colmap, furukawa2009accurate}. Although these approaches are successful, the need for many images limits their practicality in sparse-view scenarios~\cite{mildenhall2020nerf,kerbl20233dgs}.

Deep learning has significantly advanced MVS capabilities, addressing the limitations of traditional methods. Seminal works DUSt3R~\cite{dust3r_cvpr24} and MASt3R~\cite{mast3r_eccv24} accurately estimate dense 3D geometry from multi-view images, making them ideal for initializing scenes for further refinement. Building upon these models allows us to focus on enhancing \textit{geometric fidelity} rather than reconstructing scenes from scratch. However, directly refining the point clouds from MVS models using 3D Gaussian Splatting (3DGS)~\cite{kerbl20233dgs}, a differentiable rendering pipeline, can inadvertently degrade geometry, even when the initial point cloud is correctly structured. \cref{fig:teaser,fig:geom_artifact_example} illustrate geometric artifacts that arise during such refinement.
We hypothesize that these artifacts originate from Gaussians overfitting during refinement by using their excessive positional degrees of freedom (DoFs) to fit texture, creating geometric inconsistencies.

To address the issues arising from excessive DoFs, we propose a novel approach that emphasizes the importance of separating DoFs based on their inherent uncertainties. By carefully distinguishing DoFs, we categorize them into two types: image-plane-parallel DoFs and ray-aligned DoF. This distinction is crucial, as each type has unique characteristics that affect stability during refinement. Image-plane-parallel DoFs, with lower uncertainty, are directly constrained by the observed pixels, whereas ray-aligned DoFs possess higher uncertainty and require multi-view information for accurate estimation. Recognizing these distinctions allows us to effectively limit unnecessary flexibility in the model, reducing the risk of overfitting.

To explicitly manage these separated DoFs, we introduce a \textit{reprojection-based DoF separation} method that differentiates DoFs according to their respective uncertainties. In this method, we propose a \textit{bounded offset} for image-plane-parallel DoFs, limiting their movement within a pixel. A secondary benefit of our method is that it preserves valuable per-view depth information from the MVS model, which would otherwise be lost in a na\"ive refinement approach. For the ray-aligned DoFs, we propose a \textit{visibility loss} that leverages this depth information to refine the high-uncertainty DoFs through multi-view integration. By controlling each DoF with targeted constraints, our method achieves plausible geometric refinement—an aspect often overlooked in favor of rendered quality—while preserving rendering quality and minimizing geometric artifacts.

We evaluate our method on several novel view synthesis benchmarks, including Mip-NeRF 360~\cite{barron2022mipnerf360}, which features complex camera trajectories; MVImgNet~\cite{yu2023mvimgnet}, characterized by object-centric scenes and simple camera trajectories; and Tanks and Temples~\cite{knapitsch2017tanks}, which includes unbounded scenes with simple camera trajectories. The effectiveness of our method is demonstrated through quantitative analysis, including PSNR and patch-wise Pearson correlation of depth maps, as well as qualitative visualizations.

Our contributions are as follows: (1) We introduce a reprojection-based DoF separation method that separates positional degrees of freedom (DoFs) by their uncertainties to address overfitting and geometric artifacts in MVS refinement. (2) We apply tailored constraints by limiting the movement of low-uncertainty DoFs within a pixel and refining high-uncertainty DoFs using per-view depth information. (3) Our approach consistently improves geometry without compromising rendering quality, as evidenced by comprehensive quantitative and qualitative evaluations across diverse benchmarks~\cite{barron2022mipnerf360,yu2023mvimgnet,knapitsch2017tanks}.

%% file: sec/2_related_work.tex
\section{Related Work}
\subsection{Multi-View-Stereo} 
Multi-View-Stereo (MVS) aims to generate dense 3D geometry from calibrated images. 
Traditional MVS methods rely on camera parameters from Structure from Motion (SfM)~\cite{phototourism_siggraph06,schonberger2016structure} or Simultaneous Localization and Mapping (SLAM)~\cite{orb_slam2015,lsd_slam2014} and are categorized by their geometric representations: volumetric, point-cloud, or depth map-based approaches. 
Volumetric methods use voxels but incur high memory costs~\cite{seitz1999photorealistic, kutulakos2000theory, franco2005fusion, zach2007globally}. 
Point cloud-base methods~\cite{furukawa2009accurate, furukawa2010towards, fuhrmann2014mve, schonberger2016structure} densify sparse point clouds from SfM. 
Depth map-based methods~\cite{yang2003multi, zheng2014patchmatch, galliani2015massively, schonberger2016structure} estimate per-view depth maps and fuse them for 3D reconstruction. 
However, traditional MVS struggles with surfaces that have complex illumination or lack texture due to reliance on hand-crafted feature-matching algorithms~\cite{sift,surf}.

Learning-based MVS methods~\cite{wang2024learning} address these challenges by leveraging neural networks.
Among the various approaches, some use voxels to represent the scene~\cite{choy20163d, ji2017surfacenet, murez2020atlas, sun2021neuralrecon}, while others rely on depth maps to reconstruct the 3D structure~\cite{yao2018mvsnet, wang2018mvdepthnet, cheng2020deep, gu2020cascade, wang2021patchmatchnet}.
NeRF~\cite{mildenhall2020nerf} and 3D Gaussian Splatting (3DGS)~\cite{kerbl20233dgs} are applied to perform MVS, but they require a large number of images and substantial computational resources.

The latest advancements are large-scale MVS models such as DUSt3R~\cite{dust3r_cvpr24}, MASt3R~\cite{mast3r_eccv24} and PixelSplat~\cite{charatan2023pixelsplat}.
These transformer-based models, trained on extensive datasets, perform end-to-end MVS tasks efficiently, even in sparse-view conditions, handling camera calibration, relative pose estimation, and dense point cloud reconstruction.

\subsection{Sparse-View 3D Reconstruction}
3D reconstruction typically requires a large number of images~\cite{mildenhall2020nerf,kerbl20233dgs}. While neural radiance fields~\cite{mildenhall2020nerf, barron2022mipnerf360, rabby2023beyondpixels} can generate high-quality renderings from images alone, they generally need hundreds of images to learn a scene. To improve efficiency, various methods ~\cite{Niemeyer2021Regnerf,yang2023freenerf,jain2021putting,kwak2023geconerf,yang2024gaussianobject,deng2022depthsupervisednerf,wang2023sparsenerf,xiong2023sparsegs,zhu2024FSGS,li2024dngaussian} have emerged to learn 3D scenes from only a few views.

There are two main approaches for 3D reconstruction with sparse views. The first approach relies solely on the available sparse-view data, using techniques such as depth smoothness constraints~\cite{Niemeyer2021Regnerf} or masked positional encoding and occlusion regularization~\cite{yang2023freenerf} to prevent overfitting and suppress artifacts.
However, as the number of views decreases, this data-only approach may not achieve sufficient quality. To overcome this limitation, the second approach incorporates external knowledge, such as color information through normalizing flow models~\cite{Niemeyer2021Regnerf} or feature extractors to maintain visual and semantic consistency across rendered views~\cite{jain2021putting, kwak2023geconerf} or personalized diffusion models \cite{yang2024gaussianobject}.
Because most NeRF and 3DGS rely on color-based losses, they can suffer from overfitting on texture, leading to a degradation in geometry quality when the number of training views is limited~\cite{yang2023freenerf,zhu2024FSGS}.
Some studies leverage the depth of prior knowledge from pretrained models. 
For instance, depth maps generated by monocular depth estimators can act as valuable geometry priors \cite{deng2022depthsupervisednerf, wang2023sparsenerf, xiong2023sparsegs, zhu2024FSGS, li2024dngaussian}. 

\subsection{Sparse-View 3DGS with Learning-based MVS}
Unlike traditional approaches that train 3DGS initialized with sparse point clouds from SfM, \cite{yang2024gaussianobject, fan2024instantsplat} initialize Gaussians with dense point clouds from learning-based MVS, achieving high novel-view synthesis performance even with few images.
However, during training, the geometry produced by learning-based MVS models often degrades.

%% file: sec/3_preliminary.tex
\section{Preliminary}

\textbf{Learning-based MVS Methods} including DUSt3R~\cite{dust3r_cvpr24} and MASt3R~\cite{mast3r_eccv24} are learning-based multi-view stereo (MVS) methods designed to perceive 3D geometry from unconstrained multi-view images. Leveraging prior knowledge from large-scale training, they successfully reconstruct dense 3D point clouds even in sparse-view conditions, where traditional MVS methods often struggle. Building upon the architecture of DUSt3R, MASt3R incorporates an additional head for correspondence matching, enhancing its performance in estimating camera calibration, relative poses, and dense point cloud reconstruction in a single forward pass. These models leverage a transformer-based encoder-decoder architecture~\cite{transformer_NIPS2017,vit_iclr2021}, utilizing cross-attention in the decoder to capture inter-view information by processing pairs of images.

To handle multi-view geometry, DUSt3R and MASt3R simplify the problem by breaking it down into sets of two-view geometries. 
They then employ a global alignment strategy to merge pairwise predictions into a shared coordinate system. 
To lighten the optimization, multi-view geometry consistency is disregarded during global alignment. While this approach is efficient and well-suited to deep learning frameworks, it can lead to suboptimal geometry because the optimization process disregards information from other views, leading to a lack of geometric consistency. It makes further refinement necessary.

\noindent\textbf{3D Gaussian Splatting (3DGS)} is a recent differentiable rendering technique that represents scenes using 3D Gaussians as primitives. It initializes the scene with a sparse point cloud, typically obtained from Structure-from-Motion (SfM)~\cite{schonberger2016colmap} and assigns these points as the means of the Gaussians. The differentiable rasterizer in 3DGS allows for scene refinement using photometric loss, making it compatible with incorporation after an MVS module like MASt3R.

However, directly refining the point cloud from MASt3R using 3DGS can introduce geometric artifacts, such as floaters and unintended geometric patterns aligned with textures, as shown in \cref{fig:geom_artifact_example}. These issues highlight the need for a more careful integration and refinement strategy to preserve the geometric fidelity of the reconstruction.

%% file: sec/4_method.tex
\section{Method}
\label{sec:method}

We aim to reconstruct a visually and geometrically plausible scene from a limited set of images.
In \cref{sec:mvs_init}, we initialize the scene using a learning-based multi-view stereo (MVS) model. 
Next, \cref{sec:dof_separation} differentiates the Degrees of Freedom (DoFs) and proposes a reprojection-based DoF separation method that highlights their uncertainty difference.
Finally, in \cref{sec:offset_visibility_loss}, we formulate constraints for each DoF type, reducing unnecessary DoFs and enabling geometrically stable refinement even in sparse-view scenarios.

\subsection{MVS Initialization}
\label{sec:mvs_init}

We use MASt3R~\cite{mast3r_eccv24}, a variant of DUSt3R~\cite{dust3r_cvpr24}, to initialize the scene.
Compared to DUSt3R, MASt3R includes an additional head for correspondence matching, offering improved performance.
While MASt3R provides added functionality, its role in our context remains the same: it takes images as input and outputs a dense point cloud along with estimated camera poses. A formal description of MASt3R's behavior is as follows:
\begin{equation}
    \text{MASt3R}:  \{I_i\}^N_{i=1} \to \{ (D_i, T_i) \}^N_{i=1} \to \{X_i\}^N_{i=1},
\end{equation}
where $I_i$ is the input image from view $i$; $D_i$ represents the predicted per-view depth map for $I_i$; $T_i$ denotes the transformation from pixel coordinates to world coordinates, parameterized by the camera pose and intrinsic parameters; and $X_i$ represents the unprojected 3D points for view $i$, computed using the $D_i$ and $T_i$.

The top row of \cref{fig:overall_framework} shows a schematic of the behavior common to learning-based MVS models like DUSt3R and MASt3R, along with a na\"ive approach that treats MASt3R as a black-box initializer. This setup feeds the MVS point cloud into 3DGS for scene refinement.

\subsection{Reprojection-based DoF Separation}
\label{sec:dof_separation}

Refining the output of the MVS with 3DGS introduces geometric artifacts along texture as shown in \cref{fig:geom_artifact_example}. We hypothesize that this is because Gaussians leverage their excessive positional DoFs to reduce photometric loss, resulting in geometric artifacts due to overfitting. To address this, we differentiate the positional DoFs to limit the model's excessive flexibility while retaining sufficient DoFs for accurate geometric refinement.

We propose to separate each Gaussian's three positional DoFs into two distinct components: (1) \textit{image-plane-parallel DoFs} that are parallel to the image plane and (2) a \textit{ray-aligned DoF} along the pixel ray direction. Our key insight is that these DoFs exhibit different levels of uncertainty. Image-plane-parallel DoFs are directly observed in the image and thus have low uncertainty, bounded by the pixel size. In contrast, the ray-aligned DoF remains inherently ambiguous from a single view and requires multi-view information for refinement. This separation is illustrated in the left column of \cref{fig:conical_frustum}, where the possible positions of an unprojected point form a viable frustum along the pixel ray.

To make an explicit distinction during refinement, we introduce a \textit{reprojection-based DoF separation} method. This method retains per-view depth, an intermediate representation of the learning-based MVS model, and unprojects pixels to retrieve 3D Gaussians for each render. We then attach Gaussian parameters to these points for integration with the 3DGS pipeline~\cite{kerbl20233dgs}. This allows us to use the 3DGS rasterizer to reproject the Gaussians onto the image plane, effectively rendering views. 
The ray-aligned DoF corresponds to the per-view depth in this context. Image-plane-parallel DoFs are implemented by allowing small offsets $o_i$ from the pixel center when calculating a pixel ray for the unprojection. The overall pipeline of this method is highlighted by the blue box in \cref{fig:overall_framework}.

Reprojection-based DoF separation provides granular control over two distinct types of DoFs. Additionally, it retains per-view depth estimations $D_i$ from the MVS model, which would otherwise be lost in a naïve approach.
The global alignment in the MVS model removes shift-scale ambiguity~\cite{midas_tpami22} present in methods that use a monocular depth estimator for each view~\cite{darf_neurips23,scade_cvpr23}. This ambiguity-free depth information serves as a valuable prior for robust refinement. In \cref{sec:offset_visibility_loss}, we describe how this depth information is incorporated into our refinement framework, maximizing the potential of the learning-based MVS model beyond simple initialization.

\subsection{Bounded Offset \& Visibility Loss}
\label{sec:offset_visibility_loss}

\begin{figure}[t]
  \centering
   \includegraphics[width=\linewidth]{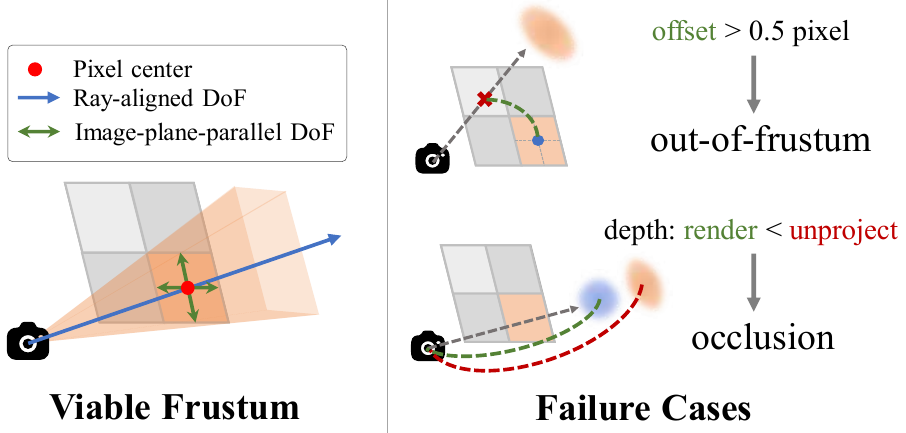}

   \caption{\textbf{Separation of DoFs.} The left column illustrates the two types of positional degrees of freedom (DoFs): ray-aligned and image-plane-parallel. The right column shows two scenarios where a point fails to project back to its original pixel.}
    \vspace{-1em}
   \label{fig:conical_frustum}
\end{figure}

To refine geometry while avoiding artifacts, we propose specific constraints for each type of DoF in our model. We elaborate on two essential conditions to ensure reprojection faithfully recovers the original views. First, the unprojected 3D point must remain within the frustum formed by rays passing through the four corners of its corresponding pixel; otherwise, it may project onto a neighboring pixel. Additionally, the point should avoid being occluded by other Gaussians in the scene, which can occur when integrating multiple views within a global coordinate system. These conditions are illustrated by the failure cases in \cref{fig:conical_frustum}.

\noindent\textbf{Bounded Offset.} For the first condition, we control image-plane-parallel DoFs by limiting the offset applied to the pixel coordinates during unprojection. Specifically, we add an offset within ±0.5 pixel units before transforming pixel coordinates into world coordinates. This ensures the 3D point stays within the viable frustum and projects back onto its original pixel. Deviating from this constraint causes the point to project incorrectly, as shown in the \textit{out-of-frustum} example in \cref{fig:conical_frustum}. The unprojection with bounded offset is formally expressed as:
\begin{align}
    \delta_i &= 0.5 \cdot \tanh(o_i), \label{eq:max_offset} \\
    X_i &= \pi^{-1}(p + \delta_i, D_i) \cdot T_i,
\end{align}
where $p$ represents the pixel center coordinates, $o_i$ denotes the learned 2D offset, $\pi^{-1}: \mathbb{R}^2 \times \mathbb{R} \to \mathbb{R}^3$ is the inverse projection function, $D_i$ is the depth map, and $T_i$ represents the camera transformation matrix. The value 0.5 in \cref{eq:max_offset} represents the maximum offset. This step corresponds to the early stage of the projection process in \cref{fig:overall_framework}, where pixel offsets are input to the unprojection function $\pi^{-1}$.

\noindent\textbf{Visibility Loss.} For the second condition, we address occlusion by defining a visibility loss. Occlusions are detected by comparing the per-view estimated depth (representing the ray-aligned DoF) with the rendered depth (which incorporates multi-view information). If the rendered depth is smaller than the per-view depth, it indicates that another Gaussian obstructs the ray, as illustrated in the \textit{occlusion} scenario of \cref{fig:conical_frustum}. The visibility loss is defined as the L2 distance between these two depths, encouraging the unprojected point to remain visible and resolving geometry conflicts introduced by multi-view alignment. Formally, the visibility loss $\mathcal{L}_{vis}$ is expressed as:
\begin{equation}
    \mathcal{L}_{vis} = \| \hat{D}_i - D_i \|^2_2,
\end{equation}
where $\hat{D}_i$ and $D_i$ denote the rendered depth and per-view depth estimated by MVS from view $i$, respectively.

%% file: sec/5_experiment.tex
\begin{table*}[ht]
\centering
\begin{tabularx}{\textwidth}{c|>{\centering\arraybackslash}X>{\centering\arraybackslash}X>{\centering\arraybackslash}X|>{\centering\arraybackslash}X|>{\centering\arraybackslash}X>{\centering\arraybackslash}X>{\centering\arraybackslash}X|>{\centering\arraybackslash}X}
\toprule
\multirow{2}{*}{Method} & \multicolumn{4}{c|}{Mip-NeRF 360~\cite{barron2022mipnerf360}} & \multicolumn{4}{c}{MVImgNet~\cite{yu2023mvimgnet}} \\ \cmidrule{2-9} 
                        & PSNR$\uparrow$ & SSIM$\uparrow$ & \multicolumn{1}{c|}{LPIPS$\downarrow$} & PDC$\uparrow$ & PSNR$\uparrow$ & SSIM$\uparrow$ & \multicolumn{1}{c|}{LPIPS$\downarrow$} & PDC$\uparrow$ \\ \midrule
InstantSplat~\cite{fan2024instantsplat} & 17.84 & 0.4468 & \multicolumn{1}{c|}{0.4359} & 0.1248 & 23.23 & \underline{0.7313} & \multicolumn{1}{c|}{0.2707} & 0.1133 \\
InstantSplat$^\dag$~\cite{fan2024instantsplat} & 17.47 & 0.4163 & \multicolumn{1}{c|}{0.4156} & 0.1528 & 23.51 & \textbf{0.7517} & \multicolumn{1}{c|}{\textbf{0.2182}} & 0.1320 \\
MASt3R~\cite{mast3r_eccv24}+3DGS~\cite{kerbl20233dgs} & \underline{18.84} & \textbf{0.4786} & \multicolumn{1}{c|}{\underline{0.4089}} & \underline{0.1537} & \underline{23.53} & 0.7154 & \multicolumn{1}{c|}{0.2555} & \underline{0.3474} \\ \midrule
Ours & \textbf{19.38} & \underline{0.4767} & \multicolumn{1}{c|}{\textbf{0.3716}} & \textbf{0.3958} & \textbf{23.57} & 0.7232 & \multicolumn{1}{c|}{\underline{0.2290}} & \textbf{0.4130} \\ \bottomrule
\end{tabularx}
\caption{
\textbf{
Quantitative comparison on Mip-NeRF 360 and MVImgNet.
}
All scenes are trained with 12 views.
PDC denotes patch-wise depth Pearson correlation. The best results are highlighted in bold, and the second-best results are underlined.
InstantSplat indicated with $\dag$ refers to the model trained for 10,000 iterations.
Our method demonstrates higher patch-wise Pearson correlation values.
}
\vspace{-0.5em}
\label{tab:res_mipnerf360_mvimgnet}
\end{table*}

\begin{table}[hbt!]
\centering
\resizebox{\linewidth}{!}{%
\begin{tabular}{c|ccc|c}
\toprule
Method & PSNR$^\uparrow$  & SSIM$\uparrow$ & LPIPS$\downarrow$ &PDC $\uparrow$            \\ \midrule
InstantSplat   & 22.22 & \underline{0.7495} & 0.1966 & \underline{0.3006}          \\
InstantSplat$^\dag$ & 21.99 & 0.7276 & \underline{0.1897} & 0.2446          \\
MASt3R+3DGS          & \underline{22.42} & {0.7451} & 0.2088 & 0.2452          \\ \midrule
Ours          & \textbf{22.80} & \textbf{0.7501} & \textbf{0.1848} & \textbf{0.6019} \\
\bottomrule
\end{tabular}%
}
\caption{
\textbf{
Quantitative comparison on Tanks and Temples.
}
All scenes are trained using 3 views.
InstantSplat indicated with $\dag$ refers to the model trained for 10,000 iterations.
Our method achieves high performance not only in geometry reconstruction but also in novel-view synthesis.
}
\vspace{-0.5em}
\label{tab:res_tnt}
\end{table}

\section{Experiment}

\subsection{Experimental Setup}
\paragraph{Datasets.}
We evaluate our method on Tanks and Temples~\cite{knapitsch2017tanks}, Mip-NeRF 360~\cite{barron2022mipnerf360}, and MVImgNet~\cite{yu2023mvimgnet}.
Mip-NeRF 360 features complex camera trajectories, MVImgNet includes object-centric scenes, and Tanks and Temples consists of unbounded forward-facing scenes. We use eight scenes from Tanks and Temples and seven from MVImgNet to align with baseline methods~\cite{kerbl20233dgs, fan2024instantsplat}.

\vspace{-1em}

\paragraph{Metrics.}
We assess 3D reconstruction quality using PSNR, SSIM~\cite{wang2004ssim}, and LPIPS~\cite{zhang2018lpips} for image fidelity. For geometry plausibility, we compare rendered depth maps with those from a monocular depth estimator~\cite{yang2024depthanything} using Pearson correlation to address scale ambiguity.
Additionally, inspired by DNGaussian~\cite{li2024dngaussian}, we compute patch-wise depth Pearson correlation (PDC) to capture local geometric details and use the average of all patches to evaluate the overall geometry of the image.
To visualize PDC, we overlay it in color on the corresponding patches of the depth map: green indicates high PDC values and plausible geometry, gray represents intermediate values, and purple signifies low PDC values, indicating implausible geometry.

\vspace{-1em}

\paragraph{Baselines.}
We use a na\"ive implementation as baseline, where we train 3DGS, initialized with the point cloud generated by MASt3R. 
Additionally, we compare with InstantSplat~\cite{fan2024instantsplat}, demonstrating state-of-the-art performance among methods using learning-based MVS. 
It enhances both the efficiency and reconstruction quality of the na\"ive approach by applying grid-based, confidence-aware Farthest Point Sampling.

\vspace{-1em}

\paragraph{Implementation Details.}
We implement our method using the gsplat~\cite{ye2024gsplatopensourcelibrarygaussian} library and trained for 10,000 iterations. We render at a resolution of 512 for the first 2,000 iterations and at the input image resolutions for the remaining 8,000. After 2,000 iterations, rendered depth maps are downsampled to match per-view resolutions for loss computation. Spherical harmonics degrees increase from 0 to 3 every 100 iterations. To prevent Gaussians from scaling excessively, we project them onto image planes and apply scale clipping at 30 pixels, following MASt3R's configuration.
For the visibility loss, we apply a linear scheduling strategy, gradually decreasing it over 10,000 iterations.

\vspace{-1em}

\paragraph{Aligning Test Views on Geometry.}
Unlike conventional methods that require highly accurate poses obtained from COLMAP~\cite{schonberger2016colmap} using hundreds of images, our approach performs reconstruction using camera poses derived from a learning-based MVS with up to 12 images.
Following NopeNeRF~\cite{bian2022nopenerf} and InstantSplat~\cite{fan2024instantsplat}, we perform 500 iterations of optimization for each test view, learning only the camera extrinsic parameters while keeping the scene frozen.

\subsection{Comparisons} 

\paragraph{Quantitative results.}
\cref{tab:res_mipnerf360_mvimgnet} shows the quantitative results for training on 12 views from the Mip-NeRF 360 and MVImgNet datasets.
The Mip-NeRF 360 dataset, with its complex camera trajectories, allows test views to reveal geometric distortions thoroughly.
In such challenging scenes, our method improves the plausibility of geometry reconstruction, resulting in a significant PSNR increase.
MVImgNet and Tanks and Temples have simpler camera parameters compared to Mip-NeRF 360.
This allows the baselines to perform novel view synthesis through simple interpolation between views \cite{bian2022nopenerf}.
However, they fail to reconstruct plausible geometry due to overfitting to texture, which is reflected in lower PDC scores.
In contrast, our method demonstrates comparable novel view synthesis performance while simultaneously reconstructing plausible geometry as shown in \cref{tab:res_mipnerf360_mvimgnet,tab:res_tnt}.
Additionally, \cref{tab:res_tnt} presents results for training on only three images from Tanks and Temples, demonstrating that our method reconstructs geometry with high fidelity, even in scenarios with very limited training images.
\vspace{-0.5em}
  
\paragraph{Qualitative results.}
\cref{fig:qualitative_result_tnt,fig:qualitative_result_mipnerf,fig:qualitative_result_mvimgnet} present the visual comparison results. 
The qualitative results show that the baselines and our method perform comparably in novel view synthesis. 
However, there is a clear difference in the quality of the rendered depth maps. 
The baseslines tend to distort geometry to represent texture.
For example, as illustrated in \cref{fig:qualitative_result_tnt}, in the Francis scene from the Tanks and Temples dataset, the ground geometry is distorted to create gaps that mimic the grid pattern in the floor texture.
The bottom row of each scene visualizes the PDC.
PDC values are represented with a color scale: patches closer to 1 are shown in green, 0 in gray, and -1 in red.
Our method demonstrates high PDC values not only for objects in object-centric scenes but also for surfaces like floors and ceilings.
This indicates that our method achieves a more plausible geometry reconstruction.

\begin{figure*}[ht]
    \centering
    \begin{minipage}[b]{0.95\textwidth}  
        \includesvg[width=\textwidth]{res/qual_res_tnt.svg}
        \caption{
        \textbf{Qualitative comparison on Tanks and Temples.}
        Our method outperforms the baselines not only in reconstructing smooth surfaces, such as the floor, but also in capturing the geometry of complex shapes like statues.
        As the color shifts from \textcolor[RGB]{87,156,44}{\textbf{green}} to \textcolor[RGB]{170,170,170}{\textbf{gray}} and then to \textcolor[RGB]{195,31,125}{\textbf{purple}}, the patch-wise depth correlation decreases, indicating less plausible geometry.
        }
        \label{fig:qualitative_result_tnt}
    \end{minipage}

    \vspace{3mm}
    
    \begin{minipage}[b]{0.95\textwidth}  
        \includesvg[width=\textwidth]{res/qual_res_mip.svg}
        \caption{
        \textbf{Qualitative comparison on Mip-NeRF 360.}
        In the Counter scene, the baselines represent the patterns of the tray and tablecloth as geometric artifacts, whereas our method more plausibly captures the geometry.
        Additionally, in the Bonsai scene, the baselines produce numerous floaters near the piano and bicycle, while our method represents the geometry without floaters.
        }
        \label{fig:qualitative_result_mipnerf} 
    \end{minipage}

    \vspace{3mm}
    
    \begin{minipage}[b]{0.95\textwidth}  
        \includesvg[width=\textwidth]{res/qual_res_mvimgnet.svg}
        \caption{
        \textbf{Qualitative comparison on MVImgNet.}
        The baselines distort the geometry to represent the texture of the surface of the chair and ground, resulting in geometric artifacts that are identifiable not only in the depth map but also in the RGB image.
        In the SUV scene, our method generates a convex geometry on the front of the SUV, whereas the baselines produce a distorted geometry.
        }
        \label{fig:qualitative_result_mvimgnet}
    \end{minipage}
\end{figure*}  

%% file: sec/6_analysis.tex
\begin{table*}[ht]
\centering
\small
\resizebox{\linewidth}{!}{%
\begin{tabular}{c|cccccc|cccccc}
\toprule
         & \multicolumn{6}{c|}{3-shots}                                                                                                                                                                         & \multicolumn{6}{c}{12-shots}                                                                                                                                                                         \\ \cmidrule(lr){2-7} \cmidrule(lr){8-13}
         & \multicolumn{2}{c}{Ours}         & \multicolumn{2}{c}{w/o Offset}                                                  & \multicolumn{2}{c|}{w/o $\mathcal{L}_{vis}$}                                    & \multicolumn{2}{c}{Ours}         & \multicolumn{2}{c}{w/o Offset}                                                  & \multicolumn{2}{c}{w/o $\mathcal{L}_{vis}$}                                     \\ 
         \cmidrule(lr){2-3} \cmidrule(lr){4-5} \cmidrule(lr){6-7} \cmidrule(lr){8-9} \cmidrule(lr){10-11} \cmidrule(lr){12-13}
         & PSNR $\uparrow$           & PDC $\uparrow$             & PSNR $\uparrow$                         & PDC $\uparrow$                            & PSNR $\uparrow$                         & PDC $\uparrow$                            & PSNR $\uparrow$           & PDC $\uparrow$             & PSNR $\uparrow$                         & PDC $\uparrow$                            & PSNR $\uparrow$                         & PDC $\uparrow$                            \\ \midrule
Ballroom & 24.01                    & \textbf{0.7766}         & 23.95                             & 0.7743                           & \textbf{24.28}                    & 0.7551                            & 30.56                    & \textbf{0.8114}         & 30.05                             & 0.8093                           & \textbf{30.64}                    & 0.7874                           \\
Barn     & \textbf{21.90}           & \textbf{0.5413}         & 21.78                             & 0.5391                           & 21.31                             & 0.5029                            & \textbf{28.33}           & \textbf{0.5931}         & 28.12                             & 0.5924                           & 27.89                             & 0.5343                           \\
Church   & \textbf{19.19}           & \textbf{0.7033}         & 19.07                             & 0.7022                           & 18.74                             & 0.6890                            & \textbf{23.55}           & 0.7597                  & 23.51                             & \textbf{0.7622}                  & 23.42                             & 0.7155                           \\
Family   & \textbf{24.11}           & \textbf{0.6552}         & 24.01                             & 0.6522                           & 23.47                             & 0.5714                            & \textbf{29.49}           & 0.6870                  & 29.20                             & \textbf{0.6873}                  & 29.42                             & 0.6076                           \\
Francis  & \textbf{23.99}           & \textbf{0.3739}         & 23.73                             & 0.3655                           & 23.54                             & 0.3356                            & \textbf{31.43}           & \textbf{0.4022}         & 30.49                             & 0.4007                           & 31.23                             & 0.3679                           \\
Horse    & \textbf{23.25}           & 0.4373                  & 23.23                             & \textbf{0.4384}                  & 23.15                             & 0.3857                            & \textbf{28.49}           & \textbf{0.4602}         & 28.23                             & 0.4592                           & 28.21                             & 0.4034                           \\
Ignatius & \textbf{23.36}           & \textbf{0.7163}         & 22.52                             & 0.7108                           & 23.32                             & 0.6684                            & \textbf{27.81}           & \textbf{0.7525}         & 26.28                             & 0.7444                           & 27.73                             & 0.7038                           \\
Museum   & \textbf{22.61}           & \textbf{0.6115}         & 22.41                             & 0.5928                           & 22.52                             & 0.5616                            & \textbf{28.63}           & \textbf{0.6729}         & 28.59                             & 0.6704                           & 28.52                             & 0.6046                           \\ \midrule
Avg.     & \textbf{22.80}           & \textbf{0.6019}         & 22.59                             & 0.5969                           & 22.54                             & 0.5587                            & \textbf{28.54}           & \textbf{0.6424}         & 28.06                             & 0.6407                           & 28.38                             & 0.5906                           \\ \bottomrule
\end{tabular}%
}
\normalsize
\caption{
\textbf{
Impact of bounded offset and visibility loss.
}
PDC denotes patch-wise Pearson correlation. The best results are highlighted in bold.
Results with 3-shot training, extremely sparse-view condition, are shown on the left, and 12-shot training on the right.
}
\label{tab:ablation}
\end{table*}

\begin{figure}[t]
  \centering
   \includesvg[width=\linewidth]{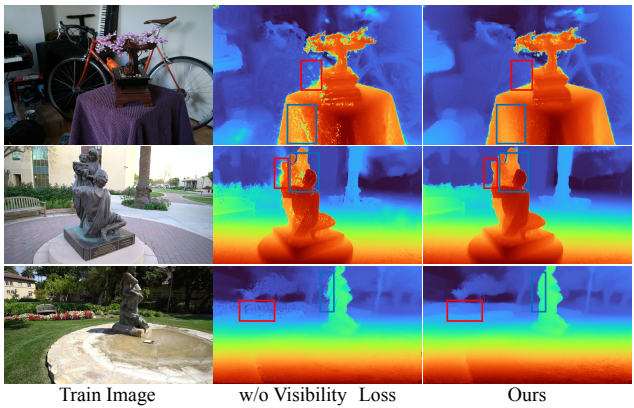}

   \caption{
   \textbf{Results of the visibility loss ablation study.} Each row presents the RGB image of a training view, the depth image without the visibility loss, and the depth image from our complete method. Omitting the visibility loss increases floaters and bumpy surfaces along textures.
   \vspace{-1em}
   \label{fig:qualitative_ablation}
}
\end{figure}

\subsection{Ablation Study}
\paragraph{Bounded Offset.} 
In \cref{tab:res_mipnerf360_mvimgnet,tab:res_tnt}, comparisons with 3DGS initialized by MASt3R demonstrate that learning the positional DoF of Gaussians with a bounded offset enables plausible geometry reconstruction. Therefore, we conduct an ablation study on learning the offset.
To evaluate the effect of learning the offset, we fix the maximum offset to zero and train on the Tanks and Temples dataset using 12 views.
As shown in \cref{tab:ablation}, when the offset is not learned, there is a marginal change in the PDC, indicating little impact on the learned geometry.
However, omitting offset training consistently degrades novel-view synthesis performance, as PSNR measures.
Conversely, in the case of 3D Gaussian Splatting initialized na\"ively with MASt3R, both PSNR and PDC exhibit low values. 
Therefore, we observe that the bounded offset contributes to the reconstruction quality in terms of both geometry and rendering quality.

\paragraph{Visibility Loss.}
We investigate the impact of visibility loss by comparing results from training on the Tanks and Temples dataset with and without its application.
\cref{tab:ablation} shows that omitting the visibility loss degrades PDC performance across all scenes, confirming that this loss contributes to learning more accurate geometry.
Improved geometry, as reflected in higher PDC, also enhances rendering quality, as evidenced by the slight drop in PSNR when the visibility loss is omitted.
This contribution is not only reflected in the PDC values but also visually in the suppression of geometrical artifacts, as shown in \cref{fig:qualitative_ablation}.

%% file: sec/7_summary_and_limiataion.tex
\section{Limitation and Future Work}
\begin{figure}[t]
  \centering
   \includesvg[width=\linewidth]{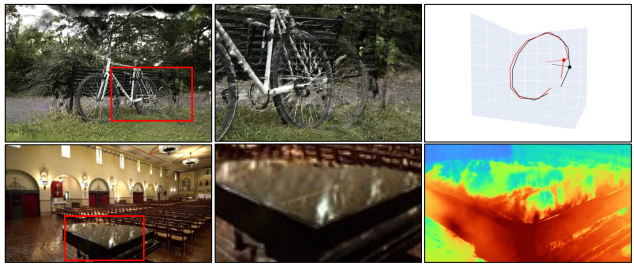}

   \caption{
   \textbf{Limitations inherited from the MASt3R.}
    The first row illustrates failures from MASt3R's inaccurate camera pose estimation, resulting in the bicycle wheel's overlapping geometries.
    The second row presents failures caused by specular surfaces, where the geometry of the piano is distorted in the reconstruction, with flat surfaces appearing hollow.
   }
   \vspace{-0.5em}
   \label{fig:limitation}
\end{figure}

We analyze cases where our method fails to reconstruct geometry plausibly and identify two major failure types inherited from MASt3R. 
First, MASt3R's error in camera pose estimation causes misalignment among unprojected per-view geometries, resulting in overlapped geometries, as shown in \cref{fig:limitation}.
Second, we observe that MASt3R interprets specular surfaces as excessively hollow, causing them to remain hollow even after refinement.
These issues could potentially cause common problems not only in our method but in all approaches that use learning-based MVS.
Therefore, future research directions could include more accurate camera pose estimation for global alignment in learning-based MVS or developing methods for accurately understanding the geometry of specular surfaces.

\section{Conclusion} 

In this work, we proposed the \textit{reprojection-based DoF separation} to improve geometry quality in sparse-view reconstruction, starting from dense geometry obtained with a learning-based MVS model.
We showed that photometric refinement without addressing excessive degrees of freedom (DoFs) leads to geometry degradation.
To address this, we separated the positional DoFs of Gaussians into image-plane-parallel and ray-aligned components based on their levels of uncertainty.
To explicitly manage each DoF, we proposed a reprojection-based DoF separation method.
Additionally, we introduced targeted constraints for DoF, considering their uncertainties, specifically bounded offset and visibility loss.
We showed that our method consistently enhances geometry quality across various dataset types.
Through this study, we showed that geometry quality is not fully captured by PSNR, and we hope that future research utilizing learning-based MVS will move beyond PSNR to also consider geometric plausibility.

\section{Acknowledgements}
This work was supported by Samsung Electronics Co., Institute of Information \& communications Technology Planning \& Evaluation (IITP) grant funded by the Korea government(MSIT) [NO.RS-2021-II211343, Artificial Intelligence Graduate School Program (Seoul National University)], the National Research Foundation of Korea (NRF) grant funded by the Korea government (MSIT) (No. 2022R1A3B1077720, 2022R1A5A708390811), and the BK21 Four program of the Education and Research Program for Future ICT Pioneers, SNU in 2024.